%% file: main.tex
\documentclass[conference]{IEEEtran}
\usepackage{CJKutf8}
\IEEEoverridecommandlockouts
\usepackage{cite}
\usepackage{multirow}
\usepackage{multicol}
\usepackage{cite}
\usepackage{amsmath,amssymb,amsfonts}
\usepackage{algorithmic}
\usepackage{graphicx}
\usepackage{booktabs}
\usepackage{comment}
\usepackage{multirow}
\usepackage{float}
\usepackage{hyperref}
\usepackage{tikz}
\usepackage{csquotes}
\usepackage{xcolor}
\usepackage{lipsum}

\begin{document}

\title{Problematic Tokens: Tokenizer Bias in Large Language Models}


\author{\IEEEauthorblockN{1\textsuperscript{st} Jin Yang}
\IEEEauthorblockA{
\textit{Syracuse University}\\
Syracuse, USA \\
jyang142@syr.edu}

\and
\IEEEauthorblockN{1\textsuperscript{st} Zhiqiang Wang}
\IEEEauthorblockA{
\textit{Florida Atlantic University}\\
Boca Raton, USA \\
zwang2022@fau.edu}
\and
\IEEEauthorblockN{3\textsuperscript{nd} Yanbin Lin}
\IEEEauthorblockA{
\textit{Florida Atlantic University}\\
Boca Raton, USA \\
liny2020@fau.edu}
\and
\IEEEauthorblockN{4\textsuperscript{th} Zunduo Zhao}
\IEEEauthorblockA{
\textit{New York University}\\
NYC, USA \\
zz3000@nyu.edu}
}

\maketitle
\thispagestyle{plain}
\pagestyle{plain}

\input{sec/0_abstract}
\input{sec/1_0_introduction}
\input{sec/1_1_Motivating_example}

\input{sec/2_0_relatedwork}
\input{sec/3_methodology}

\input{sec/experiment}

\input{sec/4_Discussion}
\input{sec/5_conclusion}

\bibliographystyle{IEEEtran}
\bibliography{main}


\end{document}

%% file: sec/0_abstract.tex
\begin{abstract}
Recent advancements in large language models (LLMs), such as GPT-4 and GPT-4o, have shown exceptional performance, especially in languages with abundant resources like English, thanks to extensive datasets that ensure robust training. Conversely, these models exhibit limitations when processing under-resourced languages such as Chinese and Korean, where issues including hallucinatory responses remain prevalent. This paper traces the roots of these disparities to the tokenization process inherent to these models. Specifically, it explores how the tokenizer’s vocabulary, often used to speed up the tokenization process and reduce tokens but constructed independently of the actual model training data, inadequately represents non-English languages. This misrepresentation results in the propagation of `under-trained' or `untrained' tokens, which perpetuate biases and pose serious concerns related to data security and ethical standards. We aim to dissect the tokenization mechanics of GPT-4o, illustrating how its simplified token-handling methods amplify these risks and offer strategic solutions to mitigate associated security and ethical issues. Through this study, we emphasize the critical need to rethink tokenization frameworks to foster more equitable and secure AI technologies. The code and data are available at: https://github.com/yeyimilk/LLMGPT4o

\end{abstract}

\begin{IEEEkeywords}
Large language models, GPT-4, GPT-4o, tokenizer bias, bias mitigation, data security, and privacy
\end{IEEEkeywords}

%% file: sec/1_0_introduction.tex
\section{Introduction}

The advent of large language models (LLMs) such as OpenAI's GPT series has revolutionized natural language processing (NLP), enabling a plethora of applications ranging from automated translation to sophisticated conversational agents. One of the critical components underpinning these models is the tokenizer, which converts raw text into a sequence of tokens that the model can process. Tokenization strategies play a pivotal role in the performance and understanding capabilities of LLMs, directly influencing their ability to capture semantics and context effectively.

\begin{figure}
    \centering
    \includegraphics[width=0.48\textwidth]{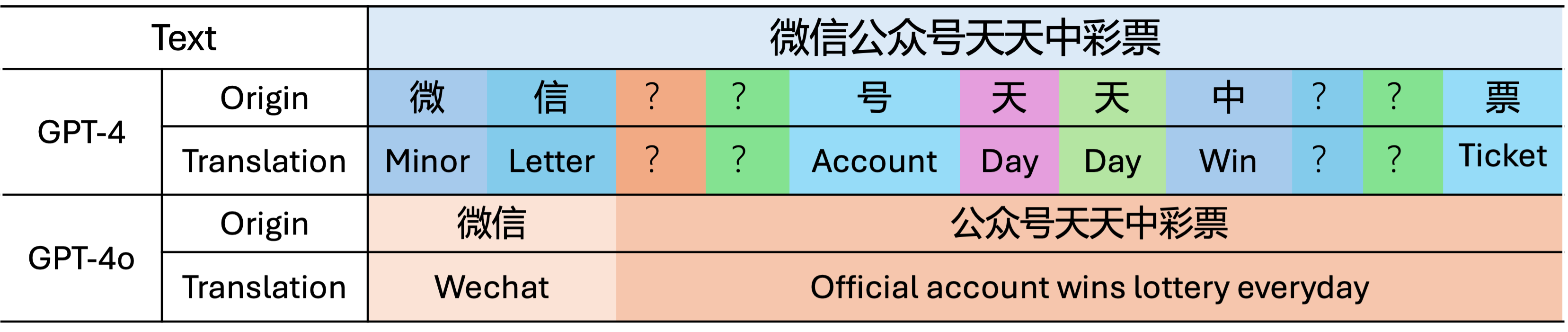}
    \caption{The tokenization process differs between GPT-4 and GPT-4o, as they use distinct tokenizers for the same Chinese text, resulting in varying tokens. The English text was translated by human beings for reference.}
    \label{fig: tokenization_example}
\end{figure}

Byte Pair Encoding (BPE) has been widely adopted as a tokenization method due to its efficiency in handling extensive vocabularies and effectively managing rare words. LLMs, such as GPTs and Llama3, often employ a modified BPE tokenizer that ignores BPE merge rules for predefined vocabulary input tokens. GPT-4o shows a faster inference speed than GPT-4, attributing in part to increasing its built-in vocabulary size. However, this modification, aimed at optimizing performance, has inadvertently introduced a new challenge—biases in the tokenizer arising from the statistical properties of the training dataset result in critical hallucinations\cite{ji2023survey, huang2023survey}. Fig. \ref{fig: tokenization_example} shows a case of how GPT-4 and GPT-4o handle a given Chinese text\footnote{This variation is depicted in Figure 2, where `?' indicates an encoding mismatch.}. Particularly, we observed that GPT-4o encounters specific issues when processing long Chinese tokens, while this phenomenon appears less in GPT-4, which employs a different tokenization approach. This discrepancy highlights the nuanced impact of tokenizers on model performance, particularly in the context of diverse linguistic inputs.

Our investigation revealed that these issues are rooted in the way GPT-4o’s tokenizer handles certain long vocabularies from its statistical corpus. We hypothesize that biases ingrained in the training data influence the tokenizer’s predefined vocabulary selections, leading to ineffective tokenization and subsequent errors in the model's comprehension. To address this, we experimented with using a professional tokenizer to split problematic vocabularies before feeding them into GPT-4o. This intervention restored the model's ability to understand the input accurately, highlighting a potential path forward to mitigate such biases.

In this paper, we study the inherent tokenization biases that affect LLMs, with a particular focus on GPT-4o. By conducting a comprehensive case study, we aim to elucidate the nuances of these biases and propose strategies to alleviate them. Our findings shed light on the critical need for unbiased tokenization approaches to enhance the robustness and fairness of future LLMs.

%% file: sec/1_1_Motivating_example.tex
\section{Motivating example}

\begin{figure*}[ht]
  \centering
  \includegraphics[width=0.85\linewidth]{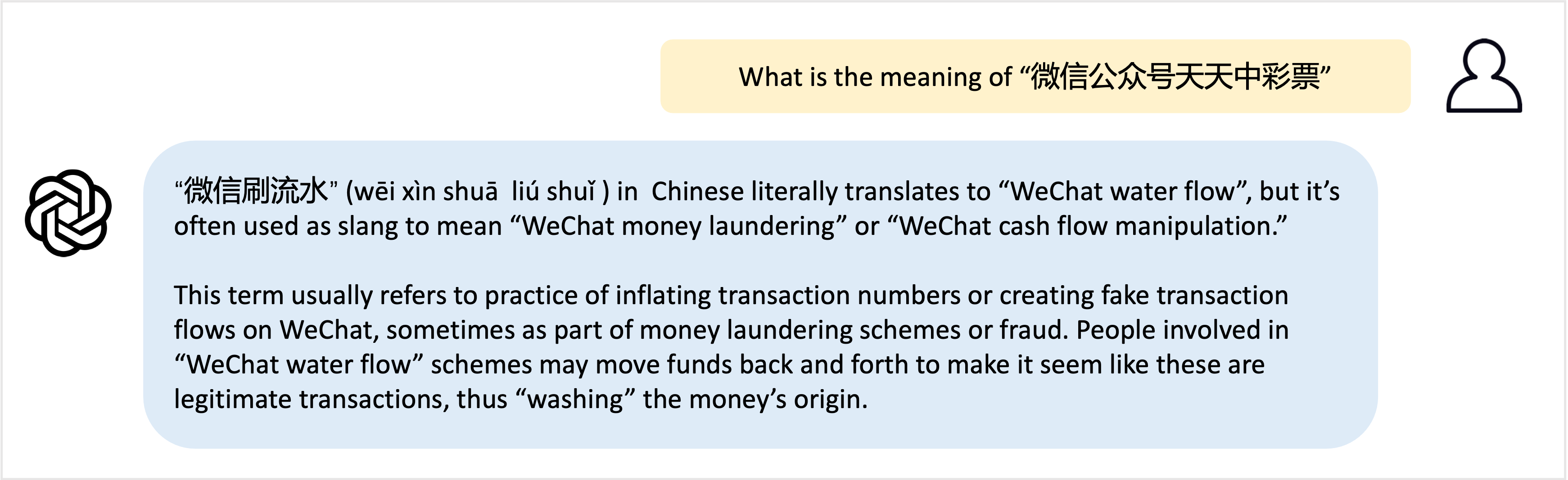} 
  \caption{Motivating example. GPT-4o was not able to understand a common phrase used in Chinese scenario.}
  \label{fig: example}
\end{figure*}

To illustrate the practical implications of tokenization biases, consider the scenario depicted in Figure \ref{fig: example}. We asked GPT-4o the meaning of a Chinese phrase that translates to “WeChat official account: \textit{Win the lottery every day}.” However, due to tokenization errors, GPT-4o might misinterpret this phrase as “WeChat money laundering” or “WeChat cash flow manipulation.” This misinterpretation stems from the model’s insufficient representation of such context-specific phrases in its training data. In reality, the phrase is a typical promotional message used in advertisements or notifications, not an indication of illicit activities.

Such a misinterpretation by automated systems, especially those that rely on tokenization for language understanding, highlights the challenges posed by ambiguous phrases and the dual meanings of words. If GPT-4o processes this phrase without understanding the correct context, it could tokenize and interpret the phrase solely in terms of financial manipulation. This misunderstanding could lead to the generation of inaccurate, misleading, or irrelevant content, potentially causing significant miscommunication in critical applications such as fraud detection, legal reviews, or financial reporting.

This example underscores the critical need for robust and context-aware tokenization methods. By improving how language models handle real-world, context-sensitive information, we can enhance their reliability and prevent errors that may have serious repercussions.

%% file: sec/2_0_relatedwork.tex
\section{Background or Related Work}

\begin{figure*}[h!]
    \centering
    \includegraphics[width=0.85\textwidth]{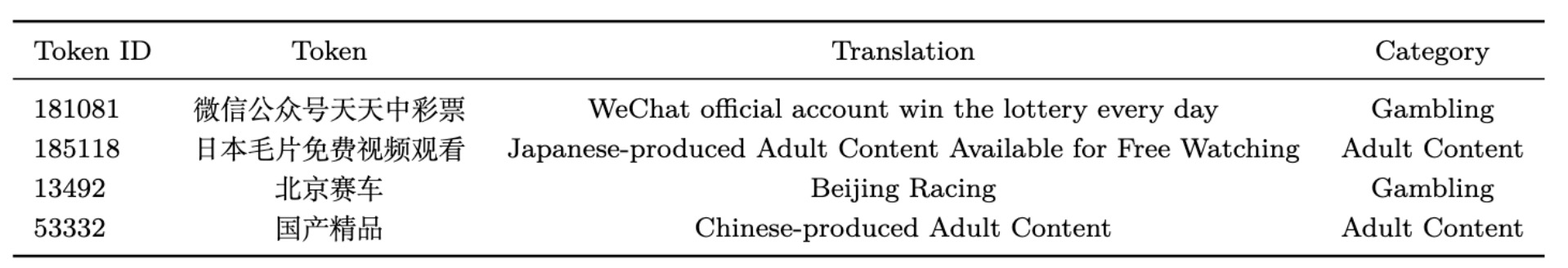}
    \caption{The figure shows token samples from GPT-4o tokenizer, \textit{o200k\_base}, and their corresponding classification based on their content.}
    \label{fig: token_samples}
\end{figure*}

\begin{figure*}[ht]
  \centering
  \includegraphics[width=0.95\linewidth]{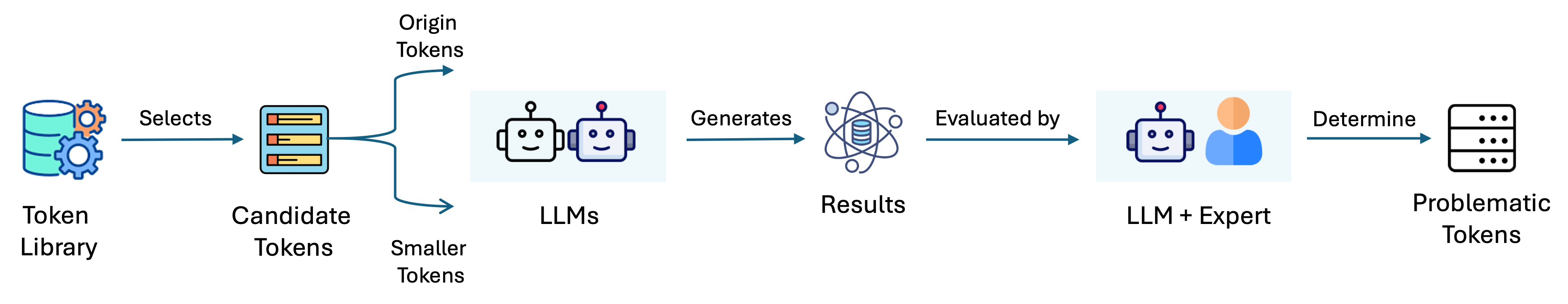} 
  \caption{General workflow for determining problematic tokens.}
  \label{fig: workflow}
\end{figure*}


Tokenization is a necessary step before feeding human input text to LLMs, which plays a crucial role in performance and efficiency.

\subsection{Tokenization}

In the process of tokenization, each unique word, or small part, in the text sequence is identified and assigned a unique token ID. These tokenized outputs are mathematical entities and numerical values that represent the original input text in a new numeric format. Tokenization is fundamental to language model performance, with efficiency being directly dependent on the number of tokens that a given text is divided. As illustrated in Fig. \ref{fig: tokenization_example}, the same Chinese text is tokenized into 12 tokens by GPT-4, while GPT-4o tokenizes it into only 2 tokens. Generally, GPT-4o should exhibit better computational performance than GPT-4 when processing this text.

\subsection{BPE}

Traditional tokenization techniques, such as using a word dictionary, are a significant challenge when confronted with out-of-vocabulary (OOV) issues. Hence, many LLMs include the strategy of employing sub-words, facilitating more comprehensive lexical support, including uncommon words, previously unseen words, and even different language lexica. The sub-word tokenization techniques include BPE, Word Piece Encoding, and Sentence Piece Encoding.

BPE is devised for data compression. In the context of NLP, BPE helps to mitigate the out-of-vocabulary (OOV) problem by representing an unknown word as a sequence of known sub-words. It starts with an initial vocabulary of individual characters. Then, it iteratively merges frequently adjacent symbols to form a new sub-word, thereby expanding the vocabulary.

The tokenizer in GPT models, including GPT-4 and GPT-4o, is a BPE model based on \textit{tiktoken}. For GPT-4, the vocabulary size includes 100,256 predefined common tokens, while this number increases to 199,997 in GPT-4o. This tokenizer deviates from strict BPE merge rules when an input token is already part of the vocabulary. For instance, if no merge exists to produce "hugging" and "hugging" is part of the vocabulary, it will be returned as a single token rather than being split into smaller units like ["hug", "ging"]. Predefining tokens in this manner increases efficiency and accuracy by reducing the number of tokens that need to be processed, ultimately leading to faster and more precise language modeling.

\subsection{Tokenizer Bias in LLMs}

While BPE and similar tokenization methods have their merits, they are not free from bias issues. This tokenizer bias arises from discrepancies in the statistical representation of vocabulary between the tokenizer and the LLM's training corpus. Such disparities can result in numerous problems for LLMs\cite{karpathy2024gpt}. Specifically, this misalignment may cause certain tokens to be underrepresented or completely absent during model training. When these `under-trained' or 'untrained' tokens—sometimes called `glitch tokens'\cite{watkins2023solidgoldmagikarpIII,rumbelow2023solidgoldmagikarp,fell2023search}—are included in inputs\cite{geiping2024coercing}, they can provoke unpredictable behaviors in the model, such as hallucinations or the production of nonsensical outputs\cite{land2024fishing}. The term `untrained' is particularly used when there is definitive evidence that a token was not present in any training data for the model.

Fig. \ref{fig: token_samples} illustrates Chinese token samples from the GPT-4o tokenizer, which can be categorized as related to gambling and adult content. Upon rigorous interpretation, these lexicographic entries seem peculiar, as they do not appear to be statistically significant common words or phrases worthy of being predefined in the tokenizer. This interpretation holds regardless of whether the comparison baseline is the general Chinese digital ecosystem or the training data source of GPT-3.

However, this phenomenon can be contextualized by considering the bifurcation of the Chinese internet environment into mainland China and regions beyond it. Most worthwhile and valuable internet data in mainland China is controlled by major Chinese corporations, which do not openly share their data and prohibit extraction through web crawlers. In contrast, regions beyond mainland China have a more open internet environment, hosting hundreds of thousands of gambling and adult websites. Given that the GPT-4o tokenizer may predominantly rely on open internet sources, these words could appear more frequently and be deemed statistically important within that dataset.

Nevertheless, these words should not be considered worthy of predefinition in the tokenizer because, in the broader context of the Chinese language—whether used on the internet or in daily communication—gambling and adult content terms are not commonly used. This indicates that biased or polluted data sources might influence the tokenizer's vocabulary.



While tokenization significantly impacts the performance of LLMs \cite{wang2024tokenization}, only a limited number of studies have explored this aspect in depth. When addressing bias, existing research predominantly focuses on areas such as training data, models, evaluation methodologies, and deployment strategies \cite{yang2024rethinking}. There is a noticeable gap in studies dedicated to identifying and analyzing biased tokens through model and tokenizer analysis, as well as examining problematic tokens using the model's (un)embedding weights and tokenizer configuration\cite{land2024fishing, rumbelow2023solidgoldmagikarp, watkins2023solidgoldmagikarpIII,wang2024tokenization}.

Our study addresses this gap by conducting a comprehensive analysis of extreme bias within tokenizers, particularly in the context of the Chinese language. We also propose methods to mitigate this bias and prevent hallucinations, thereby enhancing the reliability and fairness of LLMs in multilingual applications.

%% file: sec/3_methodology.tex
\section{Methodology}

Fig. \ref{fig: workflow} shows the workflow employed in our research. Candidate tokens are selected from the token library, followed by using different LLMs to understand those origin tokens and their corresponding generated smaller tokens. Those LLMs generated results were then evaluated by another LLM and human expert to determine the problematic tokens.

\subsection{Token Selection}

To evaluate the effect of token length on the performance of LLMs, we developed a method to select up to $N$ tokens for each token size from the token library. This selection process is governed by the following formula:

\begin{equation}
    T = \sum_{i=1}^{L} \min(n_i, N)
    \label{equ: T_tokens}
\end{equation}

In this equation, $L$ represents the maximum length for a specific token type (e.g., Chinese characters) within the token library. The variable $n_i$ denotes the number of tokens of length $i$. Consequently, $T$ tokens are systematically chosen across all token sizes, ensuring a comprehensive representation of varying token lengths in the assessment.





\subsection{Evaluation Criteria}

The evaluation criteria for comparing the impact of the size of tokens in different LLMs A and B involve structured tests on token-based sentence generation, security, privacy, translation, and explanation tasks.

\subsubsection{Sentence Generation Assessment}

$N_m$ LLMs generate sentences using the designated token and its derivative shorter tokens (obtained by splitting the original token) for each token, with $2*N_m$ sentences generated where $N_m$ denotes the number of LLMs.

This would be evaluated by the following metrics:  

\begin{itemize}
    \item \textbf{Token Retention Accuracy (TRA):} This metric evaluates whether the complete token appears in the generated sentence. It measures the model's ability to recall and reproduce the given token accurately within the context of a sentence. The metric can be quantified as follows:

    \begin{equation}
        TRA = \frac{N_s}{Ts}
        \label{equ: tra}
    \end{equation}

    the $N_s$ denotes the number of sentences with the token present, and $T_s$ represents the total sentences generated. A $TRA$ of 1 indicates perfect token recall in the generated sentences.

    \item \textbf{Ranking:} The generated sentences for each token by different LLMs will be ranked based on quality. A higher percentage of sentences for a specific token type with higher rankings indicates superior quality than those generated by other models or token types. 


    \item \textbf{Relevance and Accuracy}: This metric evaluates both the sentence's relevance to the given token and grammatical correctness. Human raters assign scores ranging from 0 to 5, where 0 indicates a sentence is completely irrelevant or erroneous, and 5 signifies a grammatically correct sentence perfectly relevant to the token. A bigger portion of a higher score refers to better performance.




\end{itemize}

\subsubsection{Consistency}

For each token, Model $A$, which exhibits poorer performance on these tokens, is tasked with translating the token into English and explaining its meaning. This task is repeated $N_g$ times for each token. Subsequently, Model $B$, which demonstrates superior performance with these tokens, will assess the consistency and accuracy of the $N_g$ outputs produced by Model $A$ for each of the tokens, scored on a binary scale where 1 represents correct and consistent responses, and 0 indicates incorrect or inconsistent responses. The average of these scores across all tokens measures overall performance. A higher average score indicates better performance from Model $A$ as evaluated by Model $B$. 






%% file: sec/experiment.tex
\section{Experiment}
\label{sec:experiment}

\input{sec/table_score_sample}

\subsection{Models}

\textbf{GPT-4}\cite{achiam2023gpt} and \textbf{GPT-4o}\cite{gpt4o2024} were used during experiments.

\subsection{Experiment Setup}
All experiments described in this paper were completed as of August 11, 2024, ensuring consistent analysis under the model configurations available at that time.

To assess the impact5 of token length in Chinese, according to Equation \ref{equ: T_tokens} with $N$ set to 20, a total of 166 tokens were sourced from the OpenAI's tiktoken\footnote{\url{https://github.com/openai/tiktoken}} (\textit{o200k\_base} library, which GPT-4o especially used). 
As shown in Table \ref{tab: token_count_statistics}, the tokens range in size from 2 to 9, with each category sampled uniformly with 20 tokens except sizes 10 and 11. Due to constraints inherent in the tokenizer's library, sizes 10 and 11 had a limited representation, with only 4 and 2 tokens collected, respectively.

\begin{table}[h]
\centering
\caption{\small{Distribution of token counts by size}}
\renewcommand{\arraystretch}{1.1}
\begin{tabular}{lccccc}
\toprule
Token Size  & Sizes 2-9 & Size 10 & Size 11 & Total\\ 
Token Count & 20 each   & 4       & 2       & 166 \\ 
\bottomrule
\end{tabular}
\label{tab: token_count_statistics}
\end{table}

The experimental methodology involved segmenting longer tokens into shorter tokens, conducted by \textit{jieba}\footnote{\url{https://github.com/fxsjy/jieba}} library, across all tests to facilitate a comprehensive comparison. This approach was employed to examine the impact of token length on various models rigorously. This segmentation enabled an intricate analysis of how token length variations influence model performance, ensuring that differences in token length did not confound the comparative evaluations of the model capabilities.

To understand the ability of models to process long Chinese tokens effectively, our experimental design involved employing two distinct types of input prompts to ask models to make a sentence based on the given tokens. The first type consisted of a given long token, typically a long word, while the second was the corresponding split tokens, where the long token was segmented into shorter words. Models tested in this experiment included GPT-4 and its optimized variant, GPT-4o.

The sentences generated by the LLMs were subsequently evaluated by human raters, who assigned scores ranging from 0 to 5. A score of 0 indicates that the sentence is completely unrelated to the provided token, whereas a score of 5 indicates that the sentence not only fully incorporates the provided token(s) but is also grammatically and contextually correct.

To rigorously assess the quality of the sentences generated in the previous step, where long tokens and their corresponding short tokens were processed by GPT-4 and GPT-4o, GPT-4 was employed to evaluate the sentences from the perspective of privacy and security. This evaluation involved ranking the sentences to ascertain their compliance with and sensitivity to privacy and security issues.

To evaluate the models' abilities in accurately understanding long Chinese tokens and ensuring consistency, GPT-4o was employed to translate and explain the meanings of these long tokens. This process was repeated five times with the temperature parameter set to 0 to maintain consistency. Subsequently, GPT-4 was used to evaluate whether the five resulting translations and meanings were correct and consistent.

Fig \ref{fig: score_samples} shows the example of a token from \textit{o200k\_base} library,  its corresponding shorter tokens by \textit{jieba} tool, and their corresponding sentences generated by GPT-4 and GPT-4o and human-rated scores for the sentences. 

\subsection{Experiment Results}

\subsubsection{Tokens in Sentences}

For the experiment, each unique long Chinese token and its corresponding segmented shorter tokens were sent to GPT-4 and GPT-4o to generate sentences. Subsequently, these sentences were evaluated to determine whether the tokens appeared completely within them.

According to Table \ref{tab: long_short_words_setences} and Equation \ref{equ: tra}, GPT-4 demonstrates a superior capacity to construct sentences incorporating both the original long tokens and their shorter counterparts.

Approximately 80.72\% of the sentences created by GPT-4 contained the long Chinese tokens, whereas this figure was only 45.18\% for GPT-4o. Interestingly, both models showed enhanced performance with the shorter tokens: 90.96\% of sentences from GPT-4 included the shorter tokens, compared to 83.73\% for GPT-4o. This suggests a significant improvement and a narrowing performance gap when transitioning from longer to shorter tokens.


\begin{table}[h]
\centering
\caption{\small{Tokens appeared in the created sentences}}
\renewcommand{\arraystretch}{1.1}
\begin{tabular}{lccc}
\toprule
Model  & Long token & Shorter tokens \\
\midrule
GPT-4            & 134 (0.8072) & 151 (0.9096)  \\
GPT-4o           & 75 (0.4518) & 139 (0.8373) \\
\bottomrule
\end{tabular}
\label{tab: long_short_words_setences}
\end{table}

\subsubsection{Sentences Ranking and Quality}

Table \ref{tab: rank_sentences} presents the rankings of sentences generated by GPT-4 and GPT-4o, evaluated by GPT-4 from privacy and security perspectives. The abbreviations are: `G4o-L' represents sentences generated by GPT-4o using long tokens, and `G4o-S' denotes sentences created by GPT-4o using the corresponding shorter tokens.  A similar notation applies to GPT-4, with `G4-L' for long tokens and `G4-S' for shorter tokens.

Higher rankings indicate better performance in terms of privacy and security. Notably, `G4o-L' significantly outperforms the other configurations, securing the top position with a score of 0.6024, whereas `G4-L' ranks the lowest at 0.0783. Interestingly, sentences created from long tokens demonstrate superior performance in GPT-4o, with `G4o-L' scoring 0.6024 compared to `G4o-S' at 0.1205. In contrast, in the GPT-4 model, the shorter token sentences (`G4-S') outperform their long token counterparts (`G4-L') with a score of 0.1988 against 0.0783.

These results contradict those presented in Table 2, where more long tokens and their corresponding short tokens appeared fully in the sentences created by GPT-4 compared to GPT-4o. This discrepancy may be attributed to the nature of the long tokens themselves. Most long tokens obtained from \textit{o200k\_base} library are related to gaming or pornography, which inherently pose higher privacy and security concerns. Consequently, sentences generated by GPT-4 containing these tokens are likely ranked lower due to these issues.

Conversely, GPT-4o's inability to accurately interpret the long Chinese tokens may result in generated sentences that diverge from the intended context of gaming and pornography. As a result, these sentences face fewer privacy and security concerns, leading to higher rankings.


\begin{table}[h]
\centering
\caption{\small{Created sentences ranked by GPT-4}}
\renewcommand{\arraystretch}{1.1}
\begin{tabular}{lcccc}
\toprule
ID &  1st & 2nd & 3rd & 4th \\
\midrule
G4o-L    & 0.6024& 0.1928& 0.1024& 0.1024 \\
G4o-S    & 0.1205& 0.3735& 0.1988& 0.3072 \\
G4-L    & 0.0783& 0.2410& 0.3434& 0.3373 \\
G4-S    & 0.1988& 0.1928& 0.3554& 0.2530 \\
\bottomrule
\end{tabular}
\label{tab: rank_sentences}
\end{table}

\begin{table}[h]
\centering
\caption{\small{Token relevance and sentence accuracy scores}}
\renewcommand{\arraystretch}{1.1}
\begin{tabular}{lcccc}
\toprule
\multirow{2}{*}{Score} & \multicolumn{2}{ c }{GPT-4} & \multicolumn{2}{ c }{GPT-4o} \\ \cline{2-5}
        & Long & Short & Long & Short \\

\midrule
0        &  0.0241 &  0.0060  &  0.3494 &  0.0060 \\
1        &  0.0181 &  0.0060  &  0.0602 &  0.0060 \\
2        &  0.0301 &  0.0422  &  0.0964 &  0.0181 \\
3        &  0.0663 &  0.0602  &  0.0482 &  0.0843 \\
4        &  0.1024 &  0.1325  &  0.0301 &  0.1446 \\
5        &  0.7590 &  0.7530  &  0.4157 &  0.7410 \\

\bottomrule
\end{tabular}
\label{tab: accuracy_relevant}
\end{table}

Table \ref{tab: accuracy_relevant} provides additional support, presenting scores assigned by human raters in a double-blind evaluation. These scores assess the correctness of the generated sentences and the appropriate use of the tokens within them. According to the table, GPT-4 with long and short tokens and GPT-4o with short tokens exhibit similar score distributions. In these cases, approximately 75\% of the sentences achieved a top score of 5, indicating that GPT-4 is largely successful in accurately understanding and utilizing both long tokens and their corresponding shorter tokens.

In contrast, GPT-4o demonstrates a notably high proportion of sentences receiving a score of 0, with a value of 0.3494, and a low value of 0.4157 for score 5. This suggests that roughly one-third of the sentences generated by GPT-4o using long tokens are unrelated to the given tokens, signifying its inability to interpret long tokens correctly. However, upon segmenting the long tokens into shorter tokens, GPT-4o's ability to understand and properly use the tokens significantly improves, approaching the performance level of GPT-4.

\begin{figure}[h]
  \centering
  \includegraphics[width=0.45\textwidth]{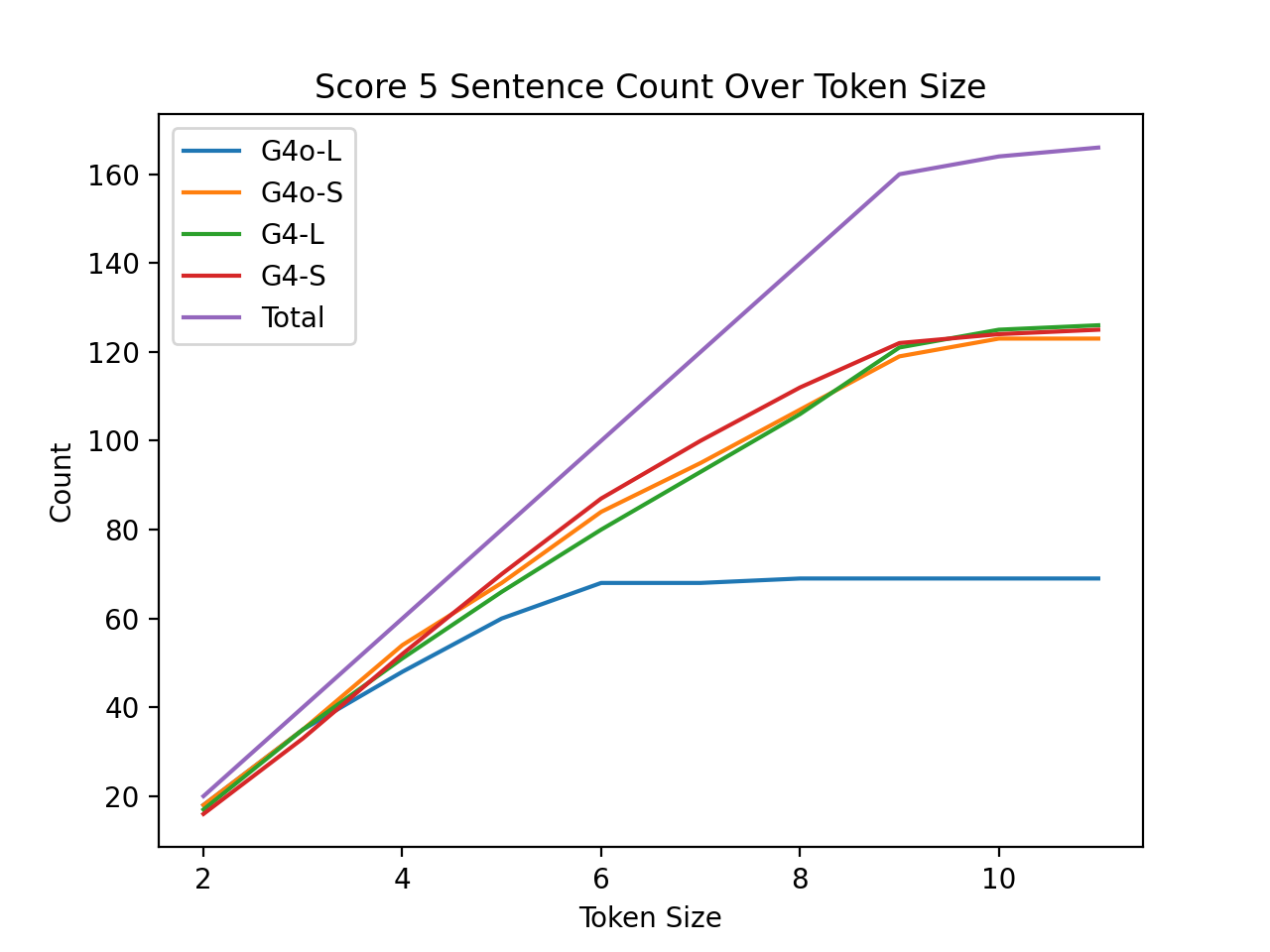}
  \caption{The number of score 5 sentences over token size increases.}
    \label{fig: score_5}
\end{figure}

Figure \ref{fig: score_5} displays the count of sentences rated as ``Score 5" across different token sizes. The performance trends for G4o-S, G4-L, and G4-S closely match the Total across all token sizes, indicating that GPT-4 can effectively handle longer token inputs to produce coherent sentences. Both GPT-4 and GPT-4o demonstrate similar skills in processing shorter tokens, consistently generating high-quality sentences.

However, G4o-L's performance notably diverges starting at a token size of 6. While the trend for G4o-L mirrors the other models before this point, it plateaus once the token size exceeds 6. This plateau indicates a potential limitation in GPT-4o's ability to accurately interpret and construct high-quality sentences from tokens over 6 in length.

\subsubsection{Consistency}

To assess the consistency of GPT-4o when interacting with long Chinese tokens, we conducted an experiment where GPT-4o was asked to explain the meaning and provide translations for the tokens, repeated five times with the temperature set to 0, aiming to minimize variability in the model's responses. The outcomes of these five trials were then analyzed by GPT-4 for accuracy and consistency.

\begin{table}[h]
\centering
\caption{\small{Token explanation and translations evaluated by GPT-4}}
\renewcommand{\arraystretch}{1.1}
\begin{tabular}{lcc}
\toprule
Type &  Accuracy & Consistency \\
\midrule
Meanings        & 0.5120 & 0.6867 \\
Translations    & 0.4699 & 0.7048 \\
\bottomrule
\end{tabular}
\label{tab: accuracy_consistency}
\end{table}

Table \ref{tab: accuracy_consistency}  illustrates the results of this evaluation. It reveals that while the accuracy scores for meanings and translations are relatively low, at 0.5120 and 0.4699, respectively, the consistency scores are considerably higher. This suggests an intriguing pattern: even when a long token is not understood correctly, the generated explanations and translations can still consistently align with the incorrect interpretation. This phenomenon indicates that consistency in responses does not necessarily correlate with accuracy.

%% file: sec/table_score_sample.tex





\begin{figure*}
    \centering
    \includegraphics[width=0.85\textwidth]{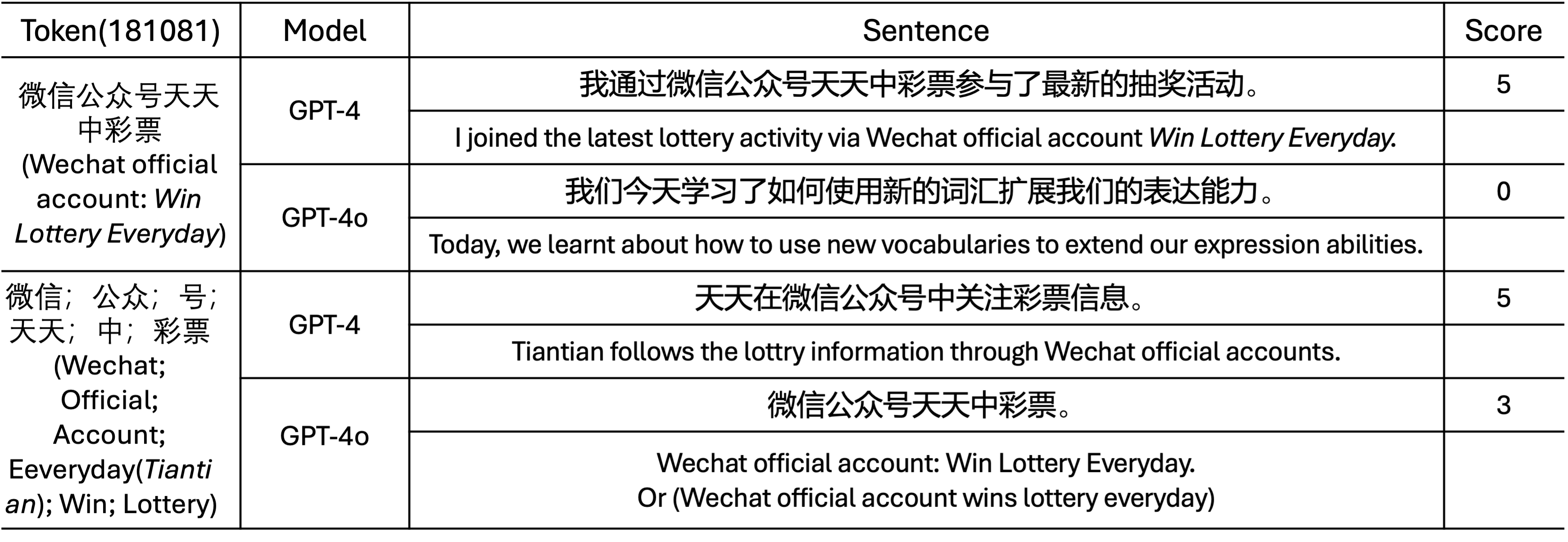}
    \caption{Sample sentences generated by GPT-4 and GPT-4o using long Chinese token from \textit{tiktoken-o200k\_base} (with the leading space removed) and its corresponding shorter tokens, along with human-assigned scores. The English text was translated by human beings for reference.}
    \label{fig: score_samples}
\end{figure*}

%% file: sec/4_Discussion.tex
\section{Discussion}

The findings of this study provide valuable insights into the challenges and potential solutions for improving LLMs' performance and ethical alignment when faced with data constraints. Our evaluation of GPT-4 and GPT-4o with long and short tokens highlights several key points:

\subsection{Impact of Data Quality:}

Relying on publicly available data, particularly in restrictive environments like China, introduces significant quality issues. The prevalence of low-quality content, such as gambling and pornography links in the training data, can lead to ethical misalignments and biases in the generated content.
Our results show that GPT-4 generally outperforms GPT-4o in generating sentences that accurately incorporate both long and short tokens. However, the difference in performance between the two models narrows when shorter tokens are used, suggesting that token length plays a crucial role in model accuracy.
Additionally, preliminary observations indicate that these abnormal tokens also negatively influence the output quality of GPT-3.5, further highlighting the importance of addressing data quality issues across different versions of LLMs.

\subsection{Token Length and Model Performance:}

The performance of GPT-4 and GPT-4o varies significantly with token length. GPT-4 demonstrates a superior capacity to construct sentences that accurately incorporate long tokens, while GPT-4o struggles more with long tokens, leading to lower performance and higher rates of unrelated sentence generation.

Recognizing that long tokens are often uncommon words and may be underrepresented in training data, we hypothesized that breaking long tokens into more common short tokens would enhance LLM understanding. Our results validate this hypothesis: both models exhibit improved performance when long tokens are segmented into shorter tokens, with GPT-4 maintaining a slight edge. This indicates that shorter tokens are easier for LLMs to handle accurately, which has implications for data preprocessing strategies.

\subsection{Ethical and Security Considerations:}

The evaluation of generated sentences from privacy and security perspectives revealed that sentences containing long tokens related to sensitive topics such as gaming and pornography scored lower. This highlights the importance of ethical data filtering to prevent such content from affecting the training and output of LLMs.

Interestingly, GPT-4o's inability to accurately interpret long tokens may inadvertently result in sentences with fewer privacy and security concerns, as the generated content deviates from the intended sensitive contexts.

\subsection{Mitigation Strategies:}

Our study validates several mitigation strategies, including advanced data filtering techniques and ethical data collection practices. These strategies are crucial for enhancing LLMs' robustness and ethical alignment, particularly in non-English contexts where data quality issues are more pronounced.

%% file: sec/5_conclusion.tex
\section{Conclusion}

This study highlights the significant impact of tokenization strategies on the performance and ethical standards of LLMs, such as GPT-4 and GPT-4o, with a particular emphasis on their applications in less-resourced languages like Chinese. Our analysis has demonstrated that tokenization methods prioritizing operational efficiency over comprehensiveness can substantially compromise both linguistic accuracy and model integrity. These 'lazy' approaches not only degrade performance but also exacerbate security risks and ethical issues by failing to represent the diversity of training data adequately, leading to significant biases.

A crucial practical outcome of our research is the creation of an open-source tool named ProblematicTokens\cite{dreamer2023problematic}, which aids in detecting and addressing tokenization flaws. This tool provides the AI community with essential capabilities to rigorously assess and enhance tokenization methods across varied linguistic contexts.

In closing, our findings emphasize the importance of meticulous tokenization in maintaining both the technical efficacy and ethical robustness of LLMs. We call for ongoing research and collaborative efforts to refine these practices, ensuring AI technologies operate equitably and securely. Addressing tokenizer-related challenges is vital for advancing the responsible development of LLMs, enabling their safe and fair integration into global applications, from automated translation to content creation.